\documentclass[letterpaper, 10 pt, conference]{ieeeconf} 
\IEEEoverridecommandlockouts
\pdfoutput=1
\usepackage{graphicx} 
\usepackage{epsfig} 
\usepackage{amsmath} 
\usepackage{amssymb} 
\usepackage{color}
\usepackage{bm}
\usepackage{subcaption}
\usepackage[linesnumbered,ruled,vlined]{algorithm2e}
\usepackage{cite}
\usepackage{makecell}
\newcommand\blue[1]{\textcolor{black}{#1}}
\newcommand{\norm}[1]{\left\lVert#1\right\rVert}

\graphicspath{{figures/}}

\title{\LARGE \bf A Joint Optimization Approach of LiDAR-Camera Fusion \\for Accurate Dense 3D Reconstructions}

\author{Weikun Zhen \thanks{Weikun Zhen and Jingfeng Liu are with the Department of Mechanical Engineering, Carnegie Mellon University,
        {\tt\small weikunz, jingfenl@andrew.cmu.edu}} \quad 
        Yaoyu Hu \quad
        Jingfeng Liu \quad 
        Sebastian Scherer\thanks{Yaoyu Hu and Sebastian Scherer are with the Robotics Institute of Carnegie Mellon University,
        {\tt\small yaoyuh,\tt\small basti@andrew.cmu.edu}}%
}
\begin{document}
\maketitle

\begin{abstract}
Fusing data from LiDAR and camera is conceptually attractive because of their complementary properties. For instance, camera images are higher resolution and have colors, while LiDAR data provide more accurate range measurements and have a ​​wider Field Of View (FOV). However, the sensor fusion problem remains challenging since it is difficult to find reliable correlations between data of very different characteristics (geometry vs. texture, sparse vs. dense). This paper proposes an offline LiDAR-camera fusion method to build dense, accurate 3D model​​s. Specifically, our method jointly solves a bundle adjustment (BA) problem and a cloud registration problem to compute camera poses and the sensor extrinsic calibration. In experiments, we show that our method can achieve an averaged accuracy of \blue{2.7}mm and resolution of 70 points/cm$^2$ by comparing to the ground truth data from a survey scanner. Furthermore, the extrinsic calibration result is discussed and shown to outperform the state-of-the-art method.
\end{abstract}\vspace{-3mm}
​\section{Introduction}
\begin{figure} [t]
    \centering
    \includegraphics[trim=0 0 0 0, clip, width=0.97\linewidth]{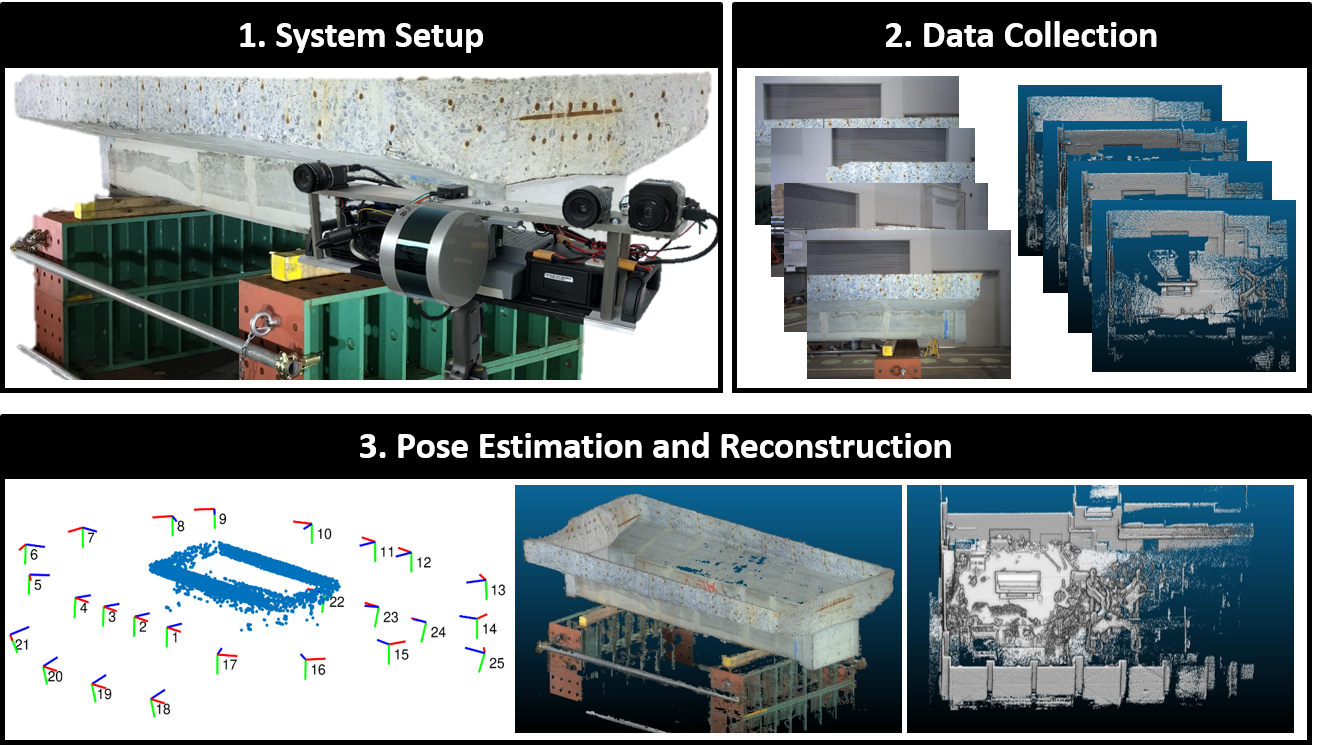}
    \caption{{A customized LiDAR-stereo system is used to collect stereo images (only left images are visualized) and LiDAR point clouds. Our algorithm estimates the camera poses, generates a textured dense 3D model of the scanned specimen and a point cloud map of the environment.}}\vspace{-3mm}
    \label{fig:head}
\end{figure}

This work is aimed at building accurate dense 3D models by fusing \blue{multiple frames of LiDAR and camera data }as shown in Fig. \ref{fig:head}. The LiDAR scans 3D points on the surface of an object and the acquired data are accurate in range and robust to low-texture conditions. However, the LiDAR data contain limited information of texture (only intensities) and are quite sparse due to the physical spacing between internal lasers. Differently, a camera provides denser texture data but does not measure distances directly. Although a stereo system measures the depth through triangulation, it may fail in regions of low-texture or repeated patterns. Those complementary properties make it very attractive to fuse LiDAR and cameras for building dense textured 3D models.

\begin{figure} [t]
    \centering
    \includegraphics[trim=0 15 0 0, clip, width=0.7\linewidth]{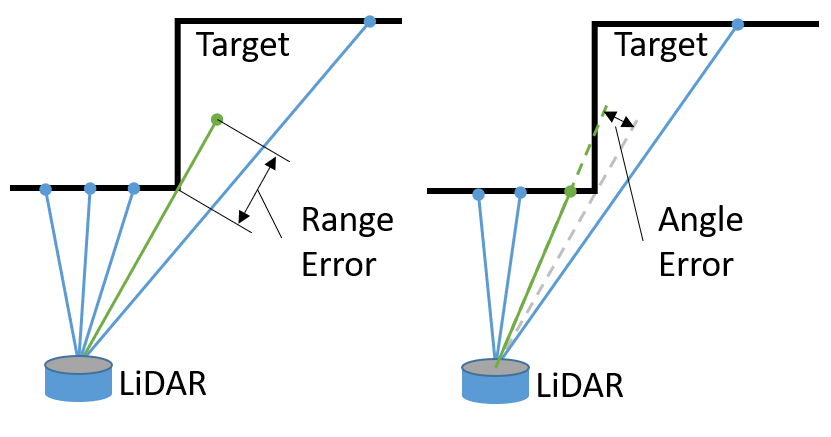}
    \caption{An illustration of inaccurate edge extraction. \textit{Left:} The \textit{mixed} edge point (green) has range error. \textit{Right:} The \textit{loose} edge point (green) has angular error.}\vspace{-4mm}
    \label{fig:best}
\end{figure}

The majority of proposed sensor fusion algorithms typically augment the image with LiDAR depth. Then the sparse depth image may be upsampled to get a dense estimation, or used to facilitate the stereo triangulation process. However, we observe two drawbacks of these strategies. The first one is that the depth augmentation requires sensor extrinsic calibration, which, compared to the calibration of stereo cameras, is less accurate since matching structural and textural features can be unreliable. For example (see Fig. \ref{fig:best}), many extrinsic calibration approaches use edges of a target as the correspondences between point clouds and images, which will have issues: 1) cloud edges due to occlusion are not clean but \textit{mixed}, and 2) edge points are not on the real edge due to data sparsity but only \textit{loosely} scattered. The second drawback is that the upsampling or LiDAR-guided stereo triangulation techniques are based on the local smoothness assumption, which becomes invalid if the original depth is too sparse. The accuracy of fused depth map is hence decreased, which may still be useful for obstacle avoidance, but not ideal for the purpose of mapping. For the reasons discussed above, we choose to combine a rotating LiDAR with a wide-baseline, high-resolution stereo system to increase the density of raw data. Moreover, we aim to fuse multiple sensor data and recover the extrinsic calibration simultaneously.

The main contribution of this paper is an offline method to process multiple frames of stereo and point cloud data and jointly optimizes the camera poses and the sensor extrinsic transform. \blue{The proposed method has benefits that:}
\begin{itemize}
\item \blue{it does not rely on unreliable correlations between structural and textural data, but only enforces the geometric constraints between sensors, which frees us from handcrafting heuristics to associate information from different domains.}

\item \blue{it joins the bundle adjustment and cloud registration problem in a probabilistic framework, which enables proper treatment of sensor uncertainties.}

\item \blue{it is capable of performing accurate self-calibration, making it practically appealing.}
\end{itemize}


The rest of this paper is organized as follows: Section \ref{sec:related} presents the related work on LiDAR-camera fusion techniques. Section \ref{approach} describes the proposed method in detail. Experimental results are shown in Section \ref{sec:experiments}. Conclusions and future work are discussed in Section \ref{sec:conclusion}.
\section{Related Work}
\label{sec:related}
In this section, we briefly summarize the related work in the areas of LiDAR-camera extrinsic calibration and fusion. For extrinsic calibration, the proposed methods can be roughly categorized according to the usage of a target. For example, a single \cite{zhou2018automatic} or multiple \cite{geiger2012automatic} chessboards can be used as planar features to be matched between the images and point clouds. Besides, people also use specialized targets, such as a box \cite{pusztai2017accurate}, a board with shaped holes \cite{velas2014calibration} or a trihedron \cite{gong20133d}, where the extracted features also include corners and edges. The usage of a target simplifies the problem but is inconvenient when a target is not available. Therefore target-free methods are developed using natural features (e.g. edges) which are usually rich in the environment. {For example, Levinson and Thrun \cite{levinson2013automatic} make use of the discontinuities of LiDAR and camera data, and refine the initial guess through a sampling-based method. This method is successfully applied on a self-driving car to track the calibration drift. Pandey et al. \cite{pandey2012automatic} develop a Mutual Information (MI) based framework that considers the discontinuities of LiDAR intensities. However, the performance of this method is dependent on the quality of intensity data, which might be poor without calibration for cheap LiDAR models. Differently, \cite{ishikawa2018lidar,schneider2013odometry,brookshire2013extrinsic} recover the extrinsic transform based on the ego-motion of individual sensors. These methods are closely related to the well-known hand-eye calibration problem \cite{tsai1989new} and do not rely on feature matching. However, the motion estimation and extrinsic calibration are solved separately and the sensor uncertainties are not considered. Instead, we construct a cost function that joins the two problems in a probabilistically consistent way and optimizes all parameters together. }

Available fusion algorithms are mostly designed for LiDAR-monocular or LiDAR-stereo systems and assume the extrinsic transform is known. For a LiDAR-monocular system, images are often augmented with the projected LiDAR depth. {The fused data can then be used for multiple tasks. For example, Dolson et al. \cite{dolson2010upsampling} upsample the range data for the purpose of safe navigation in dynamic environments. Bok et al. \cite{bok2014sensor} and Vechersky et al. \cite{vechersky2018colourising} colorize the range data using camera textures. Zhang and Singh \cite{zhang2015visual} show significant improvement on the robustness and accuracy of the visual odometry if enhanced with depth. For LiDAR-stereo systems \cite{huber2011integrating,miksik2015incremental,maddern2016real,courtois2017fusion}, LiDAR is typically used to guide the stereo matching algorithms since a depth prior could significantly reduce the disparity searching range and help to reject outliers. For instance, Miksik et al. \cite{miksik2015incremental} interpolate between LiDAR points to get a depth prior before stereo matching. Maddern and Newman \cite{maddern2016real} propose a probabilistic framework that encodes the LiDAR depth as prior knowledge and achieves real-time performance.} Additionally, in the area of surveying \cite{moussa2012automatic,neubauer2005combined,abdelhafiz2005towards}, point clouds are registered based on the motion estimated using cameras. Our method differs from these work in that LiDAR points are not projected on the image since the extrinsic transform is assumed unknown. Instead, we use LiDAR data to refine the stereo reconstruction after the calibration is recovered. 

\begin{figure*} [t]
    \centering
    \includegraphics[trim=0 0 0 0,clip,width=0.85\linewidth]{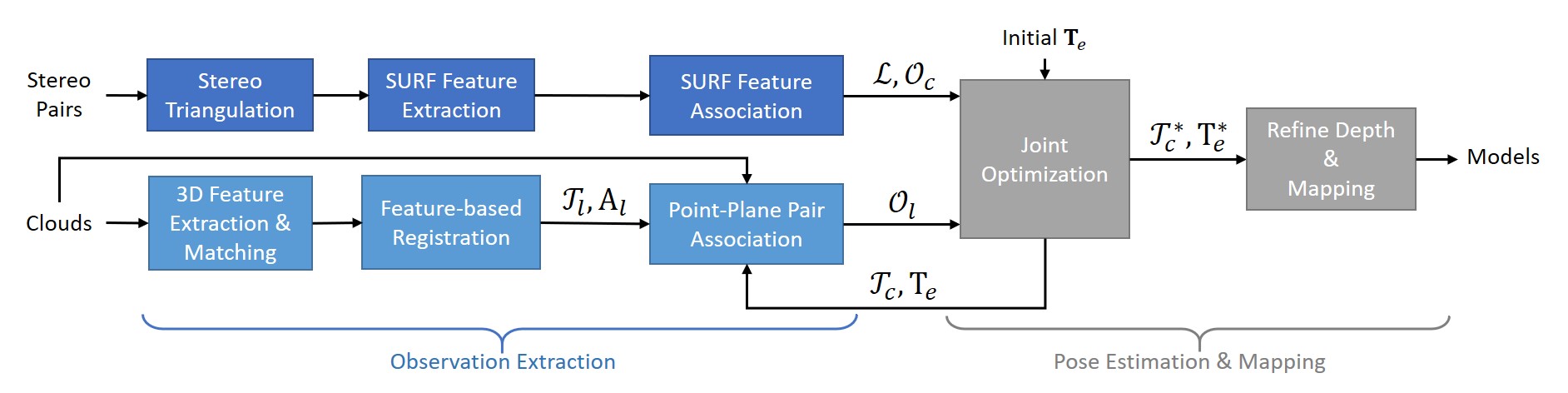}
\caption{\blue{A diagram of the proposed pipeline. In the observation extraction phase (front-end), SURF features are extracted and matched across all datasets to build the landmark set $\mathcal{L}$ and the camera observations $\mathcal{O}_c$. On the other hand, point clouds are abstracted with BSC features, and roughly registered to find cloud transforms $\mathcal{T}_l$. Then point-plane pairs are found to build the LiDAR observation set $\mathcal{O}_l$. In the pose estimation and mapping phase (back-end), we solve the BA problem and the cloud registration problem simultaneously. Here the $\mathcal{O}_l$ is recomputed after each convergence based on the latest estimation $\mathcal{T}_c,\bold T_e$ and the optimization is repeated for a few iterations. Finally, local stereo reconstructions are refined using LiDAR data and assembled to build the 3D model.}}\vspace{-4mm}
\label{fig:total_pipe}
\end{figure*}
​\section{Joint Estimation and Mapping}
\label{approach}
\subsection{Overview}
Before introducing the proposed algorithm pipeline, we clarify the definitions used throughout the rest of this paper. In terms of symbols, we use bold lower-case letters (e.g. $\bold x$) to represent vectors or tuples, and bold upper-case letters (e.g. $\bold T$) for matrices, images or maps. Additionally, calligraphic symbols are used to represent sets (e.g. $\mathcal{T}$ stands for a set of transformations). And scalars are denoted as light letters (e.g. $i, N$). 

As basic concepts, an \textit{image landmark} $\bold l \in \mathbb{R}^3$ is defined as a 3D point that is observed in at least two images. Then a \textit{camera observation} is represented by a 5-tuple $\mathbf{o}_c = \{i, k, \bold u, d, w\}$, where the elements are the camera id, the landmark id, image coordinates, the depth and a weight factor of the landmark, respectively. In addition, a \textit{LiDAR observation} is defined as a 6-tuple $\bold o_l = \{i, j, \bold p, \bold q, \bold n, w\}$ that contains the \blue{target cloud id $i$, the source cloud id $j$, a key point in the source cloud, its nearest neighbor in the target cloud}, the neighbor's normal vector and a weight factor. \blue{In other words, one LiDAR observation associates a 3D point to a local plane and the point-to-plane distance will be minimized in the later joint optimization step. }

\blue{The complete pipeline of proposed method is shown in Fig. \ref{fig:total_pipe}.} Given the stereo images and LiDAR point clouds, we first extract and match features to prepare three sets of observations, namely the landmark set $\mathcal{L}$, the camera observation set $\mathcal{O}_c$ and the LiDAR observation set $\mathcal{O}_l$. The observations are then fed to the joint optimization block to estimate optimal camera poses $\mathcal{T}_c^*$ and sensor extrinsic transform $\bold T_e^*$. \blue{Based on the latest estimation, the LiDAR observations are recomputed and the optimization is repeated. After a number of iterations, the parameters converge to local optima.} Finally, the refinement and mapping block joins the depth information from stereo images and LiDAR clouds to produce the 3D model. In the rest of this section, each component is described in detail individually. 

\subsection{Camera Observation Extraction}
Given a stereo image pair, we firstly perform stereo triangulation to obtain a disparity image using Semi-Global Matching (SGM) proposed in \cite{hirschmuller2008stereo}. The disparity image is represented in the left camera frame. Then SURF \cite{bay2006surf} features are extracted from the left image. Note that our algorithm itself does not require a particular type of feature to work. After that, a feature point is associated with depth value if a valid disparity value is found within a small radius (2 pixels in our implementation). Only the key points with depth are retained for further computation. The steps above are repeated for all stations to acquire multiple sets of features with depth. Once the depth association is done, a global feature association block is used to find correlations between all possible combinations of images. We adopt a simple matching method that incrementally adds new observations and landmarks to $\mathcal{O}_c$ and $\mathcal{L}$. Algorithm \ref{alg:img_obs} shows the detailed procedures. 
Basically, we iterate through all possible combinations to match image features \blue{based on the Euclidean distance of corresponding descriptors}. $\mathcal{L}$ and $\mathcal{O}_c$ will be updated accordingly if a valid match is found. 

\begin{algorithm}[ht]
\caption{\blue{SURF} Feature Association}
\label{alg:img_obs}
\SetAlgoLined
 Given feature sets $\mathcal{F}_{1:N}$ from $N$ stations\;
 \For{$i=1:N$}{
  \For{$j=i+1:N$}{
    \For{$\bold f$ in $\mathcal{F}_i$}{
      find the best match $\bold g$ in $\mathcal{F}_j$\;
      \If{$\bold f$ and $\bold g$ NOT similar}{
        continue\;      
      }      
      \uIf{$\bold f,\bold g$ both unlabeled}{
        create new landmark id $k \leftarrow |\mathcal{L}|$\;
        label $\bold f, \bold g$ with id $k$\;        
        add new landmark $\bold l$ with id $k$ to $\mathcal{L}$\;
        add new observations $\bold o_{\bold f}, \bold o_{\bold g}\footnotemark{}$ to $\mathcal{O}_c$\; 
      }
      \uElseIf{$\bold f$ labeled, $\bold g$ unlabeled}{
        copy label from $\bold f$ to $\bold g$\; 
        add new observation $\bold o_{\bold g}$ to $\mathcal{O}_c$\; 
      }
      \uElseIf{$\bold f$ unlabeled, $\bold g$ labeled}{
        copy label from $\bold g$ to $\bold f$\; 
        add new observation $\bold o_{\bold f}$ to $\mathcal{O}_c$\; 
      }
      \Else{
        continue\; 
      }
    }
  }
 }
 \Return $\mathcal{O}_c, \mathcal{L}$.
\end{algorithm} 
\footnotetext{$\bold o_{\bold f}, \bold o_{\bold g}$ are observation tuples filled with information from $\bold f$ and $\bold g$.}
Additionally, an adjacency matrix $\bold {A}_c$ encoding the correlation of the images can be obtained. Since the camera FOV is narrow, it is likely that the camera pose graph is not fully connected. Therefore, additional connections have to be added to the graph, which is one of the benefits of fusing point clouds. 

\begin{figure} [t]
    \centering
    \includegraphics[trim=0 0 0 0,clip,width=0.95\linewidth]{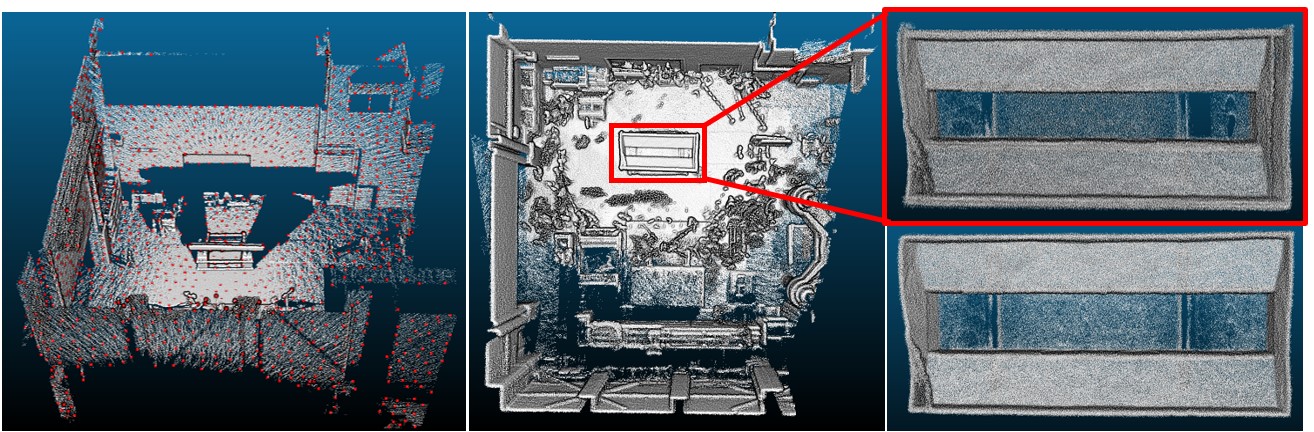}
\caption{\blue{\textit{Left:} An example of extracted BSC features (red) from a point cloud (grey). \textit{Middle:} Registered point cloud map based on matched features. \textit{Right:} Comparison of rough registration (top-right) and refined registration (bottom-right) in a zoomed-in window. }}\vspace{-4mm}
\label{fig:bsc}
\end{figure}
\subsection{LiDAR Observation Extraction}
Although many 3D local surface descriptors have been proposed (a review is given in \cite{guo2016comprehensive}), they are less stable and not accurate compared to image feature descriptors. In fact, it is preferable to use 3D descriptors for rough registration and refine the results using slower but more accurate methods such as Iterative Closest Point (ICP) \cite{besl1992method}. Our work follows a similar idea. 
Specifically, the Binary Shape Context (BSC) descriptor \cite{dong2017novel} is used to match and roughly register point clouds to compute the cloud transforms $\mathcal{T}_l$. \blue{As a 3D surface descriptor, BSC encodes the point density and distance statistics on three orthogonal projection plane around a feature point. Furthermore, it represents the local geometry as a binary string which enables fast difference comparison on modern CPUs. Fig. \ref{fig:bsc}-left shows an example of extracted BSC features. However, feature-based registration is of low accuracy. As shown in the right plots of Fig. \ref{fig:bsc}, misalignment can be observed in the rough registered map. As a comparison, the refined map of higher accuracy obtained by our method is also visualized.}

After the rough registration, another adjacency matrix $\bold A_l$ encoding matched cloud pairs is obtained. We use the merged adjacency matrix $\bold A_c \vee \bold A_l$ to define the final pose graph, where $\vee$ means element-wise or logic operation.

To obtain $\mathcal{O}_l$, a set of points are sampled randomly from each point cloud as the key points. \blue{Note that the key points to refine the registration are denser than the features.} For each pair of connected clouds in $\bold A_l$, the one with a smaller index is defined as the target while the other one as the source. Then each key point in the source is associated with its nearest neighbor and a local normal vector in the target within a given distance threshold. Finally, all point matches are formatted as a LiDAR observation and stacked into $\mathcal{O}_l$. 


\subsection{Joint Optimization}
Given the observations $\mathcal{O}_c$ and $\mathcal{O}_l$, we first formulate the observation likelihood as the product of two probabilities
\begin{equation}
P(\mathcal{O}_{c}, \mathcal{O}_{l}| \mathcal{T}, \mathcal{L},\mathbf{T}_e) = P(\mathcal{O}_{c} | \mathcal{T}, \mathcal{L}) P(\mathcal{O}_{l} | \mathcal{T},\mathbf{T}_e)
\label{eqn:likelihood}
\end{equation}
where 
$\mathcal{T}=\{\bold T_i|i=1,2,\cdots\}$ is the set of camera poses with $\bold T_1 = \bold I_4$, and $\bold T_e$ is the extrinsic transform. 
Assuming the observations are conditionally independent, we have
\begin{equation}
P(\mathcal{O}_{c} | \mathcal{T}, \mathcal{L}) = \prod_{\bold o_c\in\mathcal{O}_c} P(\bold o_c|\bold T_{i}, \bold l_{k})\label{eqn:img_prob}
\end{equation}
\begin{equation}
P(\mathcal{O}_l|\mathcal{T}, \mathbf{T}_e) = \prod_{\bold o_l \in \mathcal{O}_l} P(\bold o_l|\bold T_{i}, \bold T_{j}, \bold T_e)
\end{equation}
where $i,j$ are camera ids and $k$ is the landmark id, which are specified by observation $\bold o_c$ or $\bold o_l$. The probability of one observation is approximated with a Gaussian distribution as
\begin{align}
&P(\bold o_c|\bold T_{i}, \bold l_{k}) \propto \exp \left(-\frac{1}{2}w_{\bold o_c}(E_f^2+E_d^2)\right)\\
&P(\bold o_l|\bold T_{i},\bold T{j}, \bold T_e) \propto \exp \left( -\frac{1}{2}w_{\bold o_l}E_l^2\right)
\end{align}
where $w_{\bold o_c}, w_{\bold o_l}$ are the weighting factors of camera and LiDAR observations. \blue{And the residual $E_f$ and $E_d$ encode landmark reprojection and depth error, while $E_l$ denotes the point-to-plane distance error. Those residuals are defined as}
\begin{align}
\text{feature:}\;E_f &= \frac{||\phi(\bold l_k|\bold K, \bold T_i) - \bold u||}{\sigma_p} \label{eqn:proj_residual}\\
\text{depth:}\;E_d &= \frac{\norm{\psi(\bold l_k|\bold T_i)} - d}{\sigma_d}\\
\text{laser:}\;E_l &= \frac{\bold n^\text{T}\left(\psi(\bold p|\bold T_{l,ij}) - \bold q\right)}{\sigma_l}
\label{eqn:residual}
\end{align}
Here, $\bold u$ and $d$ are observed image coordinates and depth of landmark $k$. $\bold T_{l,ij} = (\bold T_e\bold T_i)^{-1}\bold T_j \bold T_e$ is the transform from target cloud $i$ to source cloud $j$. Function $\phi(\cdot)$ projects a landmark onto the image $i$ specified by input intrinsic matrix $\bold K$ and transform $\bold T_i$. Function $\psi(\cdot)$ transforms a \blue{3D point} using the input transformation. $\sigma_p$, $\sigma_d$ and $\sigma_l$ denote the measurement uncertainties of extracted features, stereo depths and LiDAR ranges, respectively.

Substituting (\ref{eqn:img_prob})-(\ref{eqn:residual}) back into (\ref{eqn:likelihood}) and taking the negative log-likelihood gives the cost function
\begin{equation}
f(\mathcal{T}, \mathcal{L},\mathbf{T}_e) = \frac{1}{2}\sum_{\bold o_c} w_{\bold o_c}\left( E_f^2+E_d^2\right) + \frac{1}{2}\sum_{\bold o_l} w_{\bold o_l}E_l^2
\label{eqn:cost}
\end{equation}
which is iteratively solved over parameters $\mathcal{T}, \mathcal{L},\mathbf{T}_e$ using the Levenberg-Marquardt algorithm.

\blue{To filter out incorrect observations in both images and point clouds, we check the reprojection error $||\phi(\bold l_k|\bold K, \bold T_i) - \bold u||$ and depth error $\norm{\psi(\bold l_k|\bold T_i)} - d$ of camera observations and check the distance error $\bold n^\text{T}\left(\psi(\bold p|\bold T_{l,ij}) - \bold q\right)$ of LiDAR observations after the optimization converges. The observations whose errors are larger than prespecified thresholds will be marked as outliers and assigned with zero weights. The cost function (\ref{eqn:cost}) is optimized repeatedly until no more outliers can be detected. The thresholds can be tuned by hand and in the experiments we use 3 pixels, 0.01m and 0.1m respectively. }

\blue{Similar to the ICP algorithm, the $\mathcal{O}_l$ is recomputed based on the latest estimation of $\mathcal{T}_c,\bold T_e$, while the $\mathcal{O}_c$ remains unchanged. Once $\mathcal{O}_l$ is updated, the outlier detection and optimization steps are repeated as mentioned above. The $\mathcal{O}_l$ only needs to be recomputed a few times (4 times in our experiments) to achieve good accuracy. } 

Additionally, the strategy of specifying the uncertainty parameters is as follows. Based on the stereo configuration, the triangulation depth error $e_d$ is related to the stereo matching error $e_p$ by a scale factor as in
$e_d = (d^2/bf)e_p$, where $b$ is the baseline, $f$ is the focal length and $d$ is the depth. 
Assuming the uncertainties of feature matching and stereo matching are equivalent, we have $\sigma_d = (d^2/bf)\sigma_p$. Therefore, we can now set $\sigma_p$ to be the identity (i.e. 1) and set $\sigma_d$ by multiplying the scale factor. On the other hand, the value of $\sigma_l$ is tuned by hand \blue{so that the total cost of camera and LiDAR observations are roughly at the same magnitude. In the experiments, setting $\sigma_p = 1$, $\sigma_d=5\times 10^3$ and $\sigma_l$ between $[0.02,0.1]$ can generate sufficiently good results. } 

\subsection{Mapping}
\begin{figure} [t]
    \centering
    \includegraphics[width=0.98\linewidth]{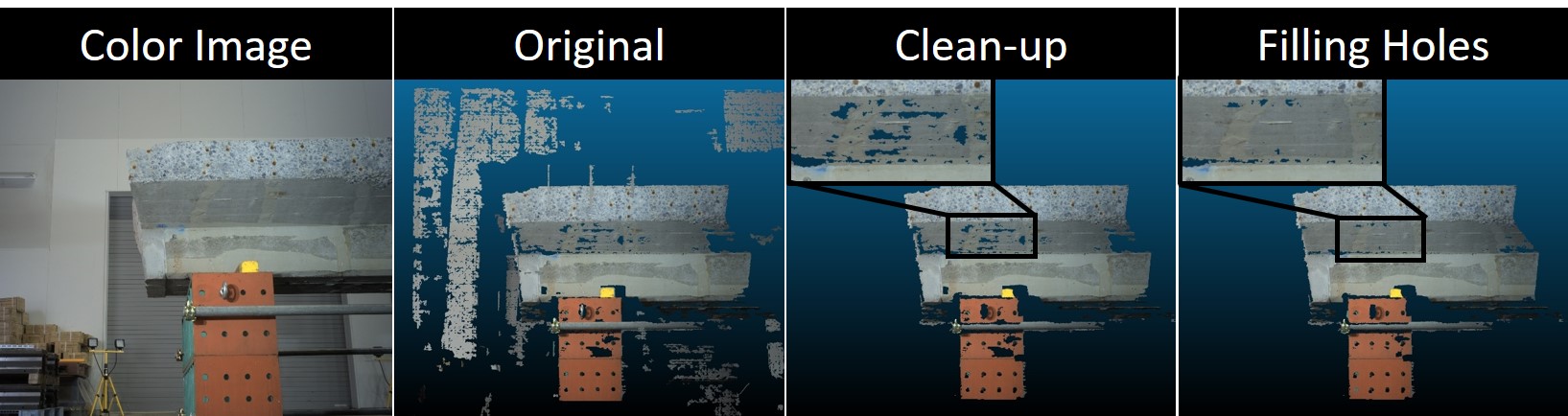}
\caption{{An example of refining the stereo depth. The outliers are first filtered out by limiting its difference to the LiDAR depth within a maximum range threshold. Then the holes are filled with the surrounding LiDAR depth only if the local surface has a near-zero curvature. }}\vspace{-5mm}
\label{fig:clean_compare}
\end{figure}

{With the camera poses estimated, building a final 3D model could be simply registering all stereo point clouds together. However, the stereo depth maps typically contain outliers and holes due to triangulation failure. In order to refine the stereo depth maps, we further perform a simple but effective two-fold fusion of LiDAR and camera data for each frame or station. In the first fold, the stereo depth is compared with the projected LiDAR depth and will be removed if \blue{there} is a significant difference. In the second fold, LiDAR depth is selectively used to fill holes in the stereo depth. Particularly, we only use the regions that are locally flat (such that the local smoothness assumption is valid), and well observed (avoiding degenerated view angle). The curvature of the local surface is used to measure the flatness. And the normal vector is used to compute the view angle. Fig. \ref{fig:clean_compare} shows an example of refining the stereo point cloud. It can be observed that holes lying on a flat surface can be filled successfully, while the missing points close to the edges are not treated to avoid introducing new outliers. 
}

\subsection{Conditions of Uniqueness}
\blue{The proposed approach relies on the ego-motion of individual sensors to recover the extrinsic transform $\bold T_e$, making it possible that $\bold T_e$ is not fully observable if the motion degenerates. It turns out to be the same problem encountered in hand-eye calibration, where the extrinsic transform between a gripper and a camera is estimated from two motion sequences. Here we discuss conditions for a fully observable $\bold T_e$ by borrowing knowledge from the hand-eye calibration, whose classical formulation is given by}\vspace{-0mm} 
\begin{equation}
\bold T_c \bold T_e = \bold T_e \bold T_h
\vspace{-1mm}
\label{eqn:classic}
\end{equation}
where $\bold T_h, \bold T_c$ represent the relative motion of the hand and the camera w.r.t. their own original frames. Incorporating multiple stations will result in a set of (\ref{eqn:classic}) and then $\bold T_e$ can be solved. According to \cite{tsai1989new}, the following two conditions must be satisfied to guarantee a unique solution of $\bold T_e$:

\begin{enumerate}
\item At least 2 motion pairs $(\bold T_c, \bold T_h)$ are observed. Equivalently, at least 3 stations are needed, with one of them to be the base station. 
\item The rotation axes of $\bold T_c$ are not colinear for different motion pairs.
\end{enumerate}
In our case, the robot hand frame is substituted by the LiDAR frame. Therefore, the configuration of each station must also satisfy the above conditions of uniqueness. {This provides formal guidance to collect data effectively. From our experience of deploying the developed system, an operator without adequate background knowledge in computer vision, particularly in structure from motion, is likely to miss the second condition and only rotates the sensor about the vertical axis, which will make the extrinsic calibration unobservable.} 
\begin{figure} [t]
    \centering
    \includegraphics[trim=0 15 0 11, clip, width=0.75\linewidth]{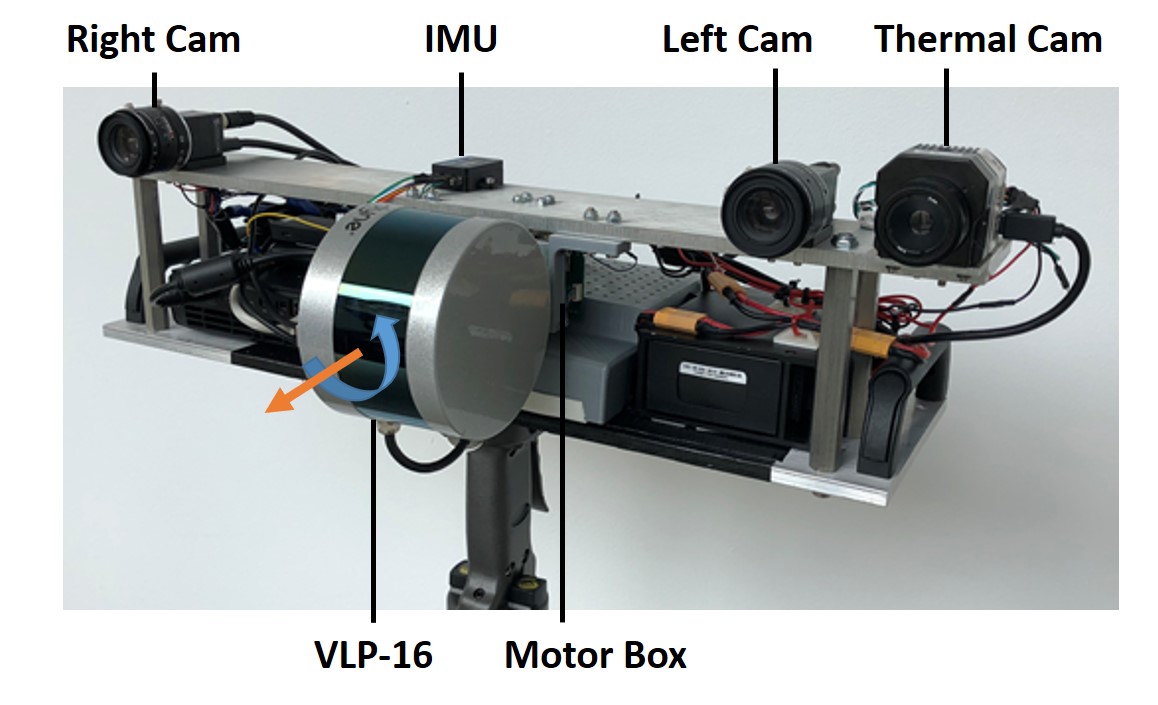}
\caption{The sensor pod developed for data collection.} \vspace{-2mm}
\label{fig:sensorpod}
\end{figure}
​\section{Experiments}
\label{sec:experiments}
\subsection{The Sensor Pod}
To collect data for experiments, we developed a sensor pod (as shown in Fig. \ref{fig:sensorpod}) which has a pair of stereo cameras (global shutter, resolution $4112\times 3008$, baseline 38cm), a Velodyne Puck (VLP-16), an IMU and a thermal camera. This work only uses the stereo image pairs and LiDAR clouds for reconstruction. Particularly, the VLP-16 is mounted on a continuously rotating ($180^\circ$ per second) motor to increase the sensor FOV. 

\blue{The calibration between the involved sensors are performed separately. We use the OpenCV library \cite{opencv_library} to obtain camera intrinsic and extrinsic parameters. The transform between the motor and the LiDAR frame is obtained by placing the sensor pod in a conference room, and carefully tuning the transform until the accumulated points on walls and ceiling form thin surfaces in the fixed motor base frame. From now on, we use the term \textit{LiDAR frame} to denote the fixed motor base frame instead of the actual rotating Velodyne frame, and assume all point clouds have been transformed into the LiDAR frame.}
\subsection{Reconstruction Tests}
\begin{figure} [t]
    \centering
    \includegraphics[trim=0 10 0 0,clip,width=0.8\linewidth]{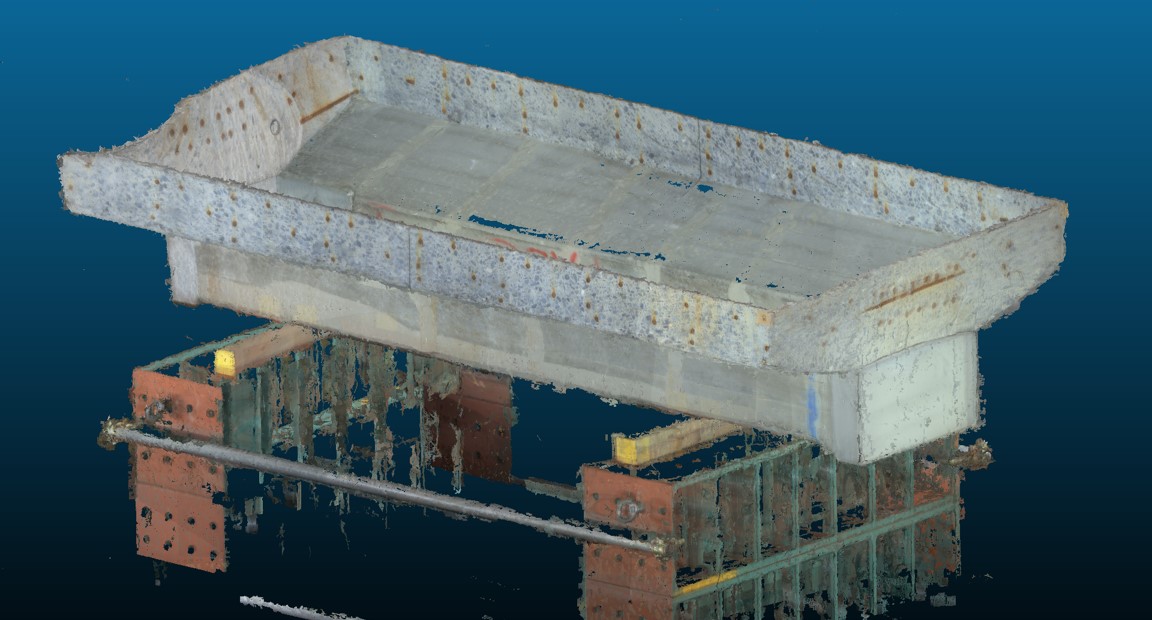}
\caption{Built point cloud model of the T-shaped specimen.}\vspace{-3mm}
\label{fig:model}
\end{figure}
\begin{figure} [t]
    \centering
	\includegraphics[width=0.7 \linewidth]{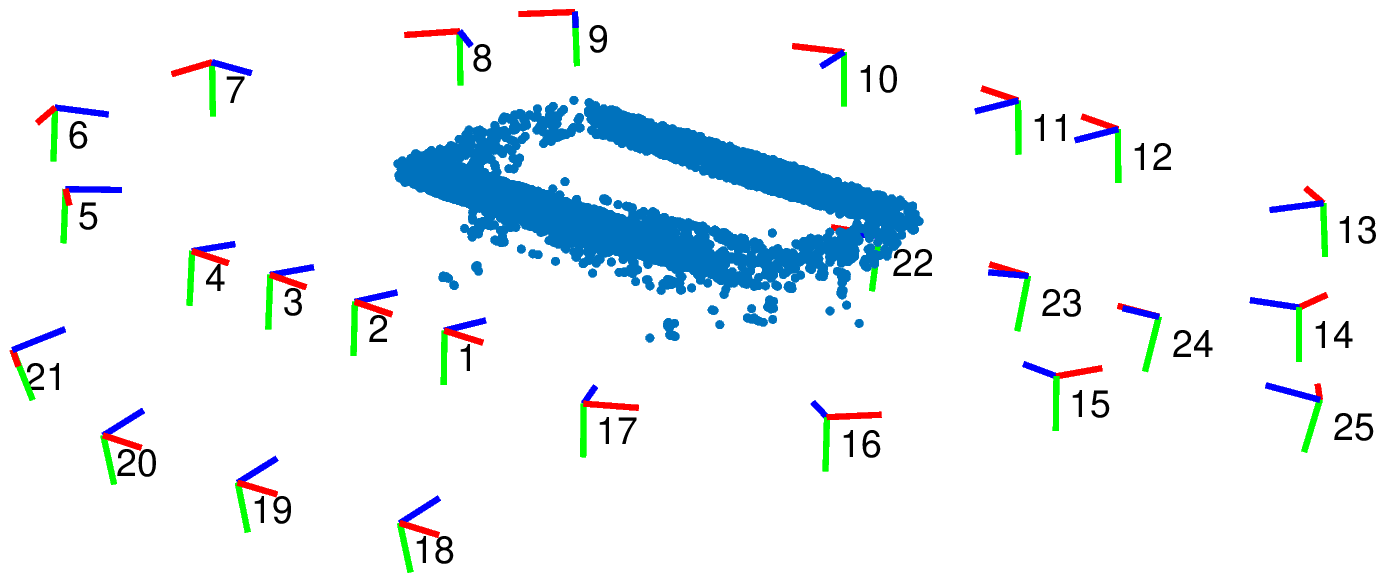}
    \includegraphics[width=0.48\linewidth]{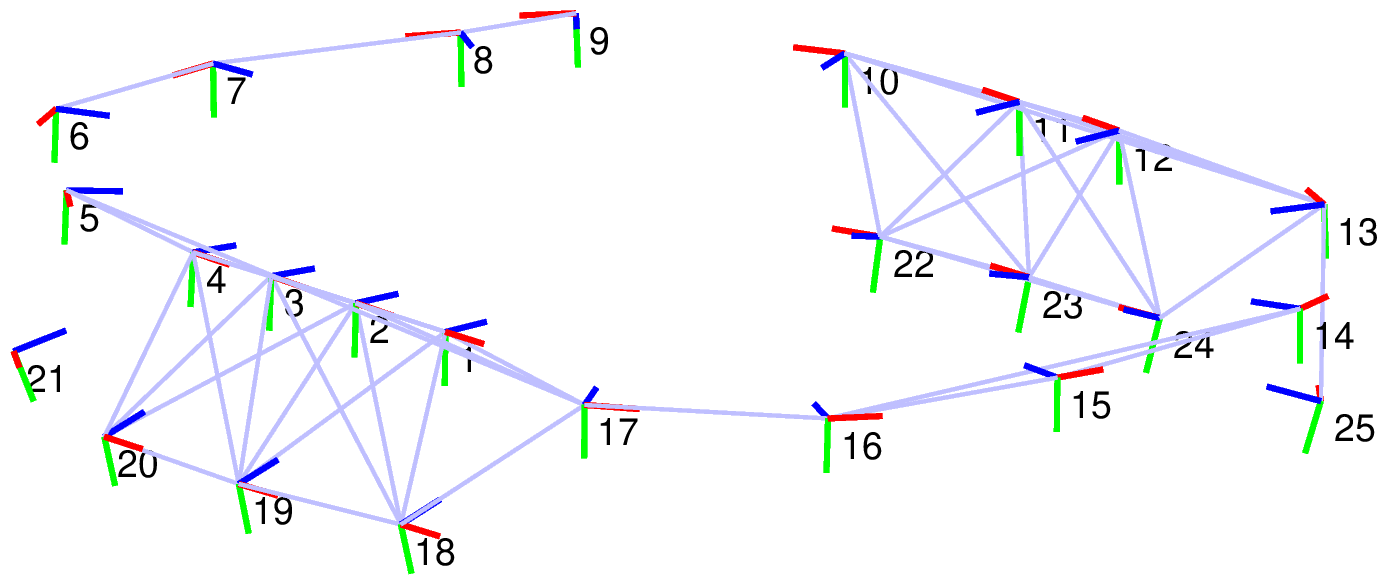}
    \includegraphics[width=0.48\linewidth]{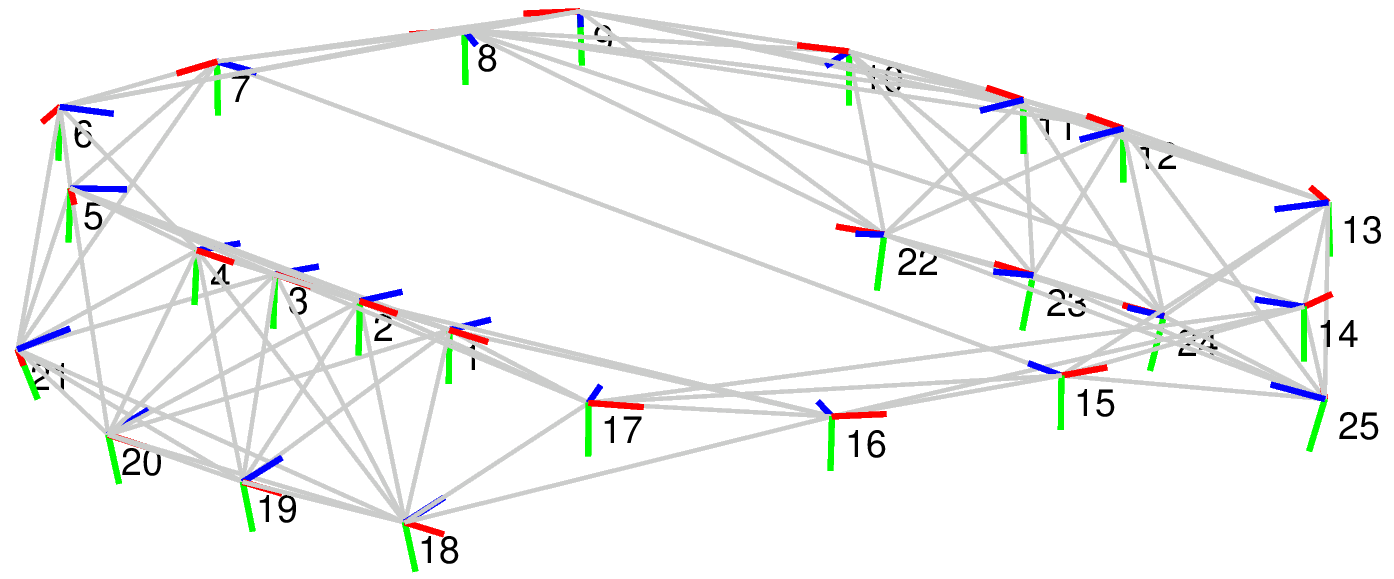}
\caption{\textit{Top:} Estimated camera poses (numbered in the order of capture) and visual landmarks (blue points). We follow the convention to define camera frame $z$ (blue) forward, $y$ (green) downward. \textit{Bottom:} Pose graph connections from images (blue) and poing clouds (gray)}\vspace{-4mm}
\label{fig:poses}
\end{figure}
\begin{figure*} [t]
    \centering
    \includegraphics[width=0.99\textwidth]{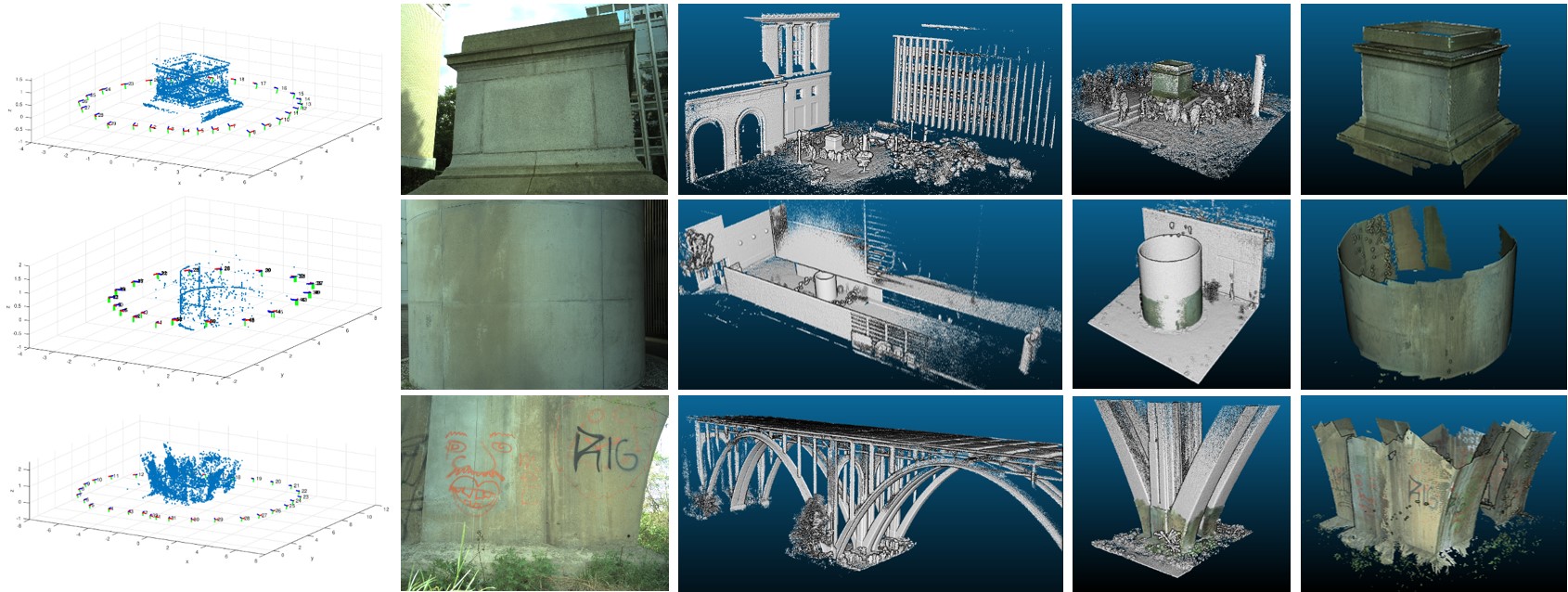}
\caption{\blue{From top to bottom, the results of three tests are visualized: a squared pillar (top), a cylinder pillar (middle) and a bridge pillar (bottom). From left to right, we visualize the camera poses and landmarks (blue points), a sample of the image data, complete LiDAR point cloud, overlaid LiDAR and stereo point cloud, dense stereo point cloud. }}\vspace{-4mm}
\label{fig:data}
\end{figure*}
The first reconstruction test is carried out at the Shimizu Institute of Technology in Tokyo to scan a T-shaped concrete specimen that is under structural tests. 
In total, 25 stations of data are collected around the specimen at a distance of about 2.5 meters. Each station contains a stereo image pair, a point cloud that accumulates scans for 20 seconds and contains approximately 1.6 million points. For station 1-17, the sensor pod is placed on a tripod and pointed to the specimen. Station 18-25 are collected with the sensor pod on the ground, tilted up to capture the bottom of the specimen. Fig. \ref{fig:model} shows the reconstructed model and Fig. \ref{fig:poses} visualizes the camera poses and landmarks. In the lower plots of Fig. \ref{fig:poses}, correlations found between images (blue lines) and point clouds (grey lines) are visualized. Since the cameras have narrow FOV ($48^\circ$ horizontal), it is likely that adjacent images don't have enough overlap, which makes the pose graph not fully connected. Fortunately, LiDAR clouds have much wider FOV and therefore guarantees a fully connected graph.


As to the computation statistics, we provide a rough measure of the processing time of the major components. On a standard desktop (i7-3770 CPU, 3.40GHz$\times$8), it takes less than 2min to remove vignetting effects and triangulate a stereo pair (40-50min for the whole dataset). The feature-based cloud registration takes about 15min in total and the joint pose estimation and map refinement can be finished in about 15min and 20min respectively.

\blue{In addition to the T-shaped specimen, we tested our algorithm in different environments, where the shapes of reconstructed objects vary from simple squared and cylinder pillars to more complex bridge pillars (see Fig. \ref{fig:data}). Table \ref{tbl:data} summarizes the model statistics. The averaged error is obtained by comparing to a ground truth model and more details are provided in Section \ref{sec:experiments}-E.} \vspace{-1mm}
\begin{table}[h]
\caption{Dataset and Model Statistics}\vspace{-2mm}
\label{tbl:data}
\begin{center}
\begin{tabular}{|c||c|c|c|c|}
\hline
Datasets & \makecell{Stations\\(Frames)} & \makecell{\# of LiDAR \\points ($\times10^6$)} & \makecell{\# of stereo \\points ($\times10^6$)} & \makecell{Error\\(mm)}\\
\hline
T-shaped & 25 & 32.4 &78.4 & N/A\\
squared& 29 & 39.1 & 210.3 & 2.7\\
cylinder& 54 & 66.5 & 111.7 & N/A\\
bridge& 32 & 38.6 & 168.7 & 3.9\\
\hline
\end{tabular}
\end{center}
\end{table}\vspace{-3mm}
\subsection{LiDAR-Camera Calibration}
\begin{figure} [t]
    \centering
    \begin{subfigure}[b]{0.12\textwidth}
      \includegraphics[width=\textwidth]{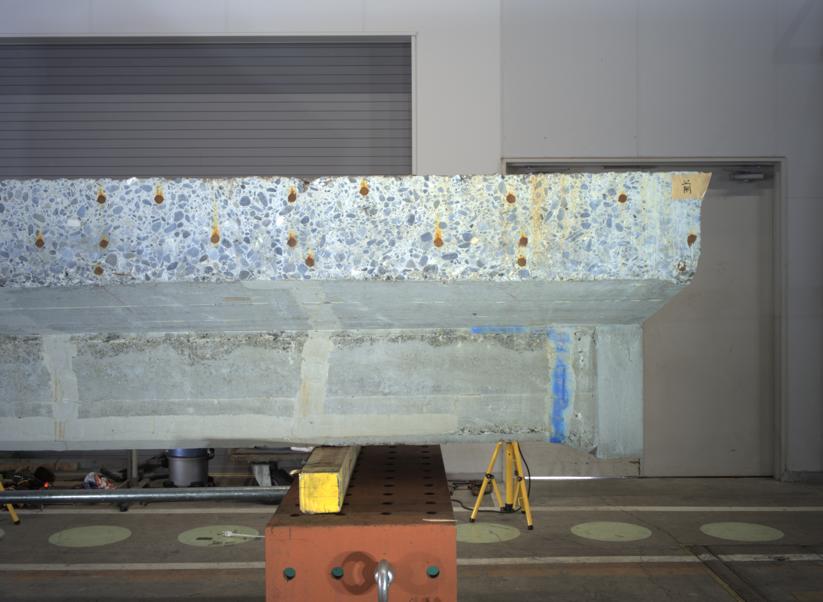}
      \subcaption{Raw image}
       \label{fig:raw_color}
    \end{subfigure}\begin{subfigure}[b]{0.12\textwidth}
      \includegraphics[width=\textwidth]{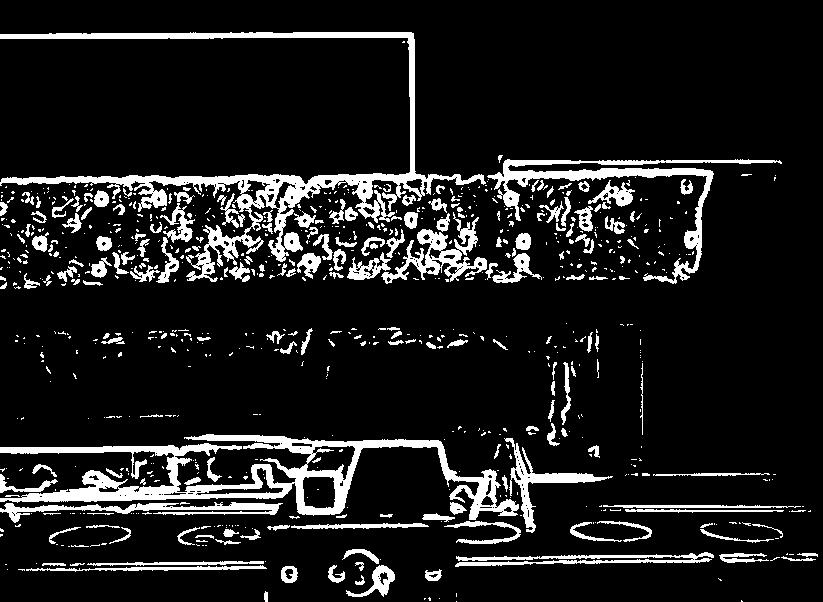}
      \subcaption{Edges}    
       \label{fig:edge_map}
    \end{subfigure}\begin{subfigure}[b]{0.12\textwidth}
      \includegraphics[width=\textwidth]{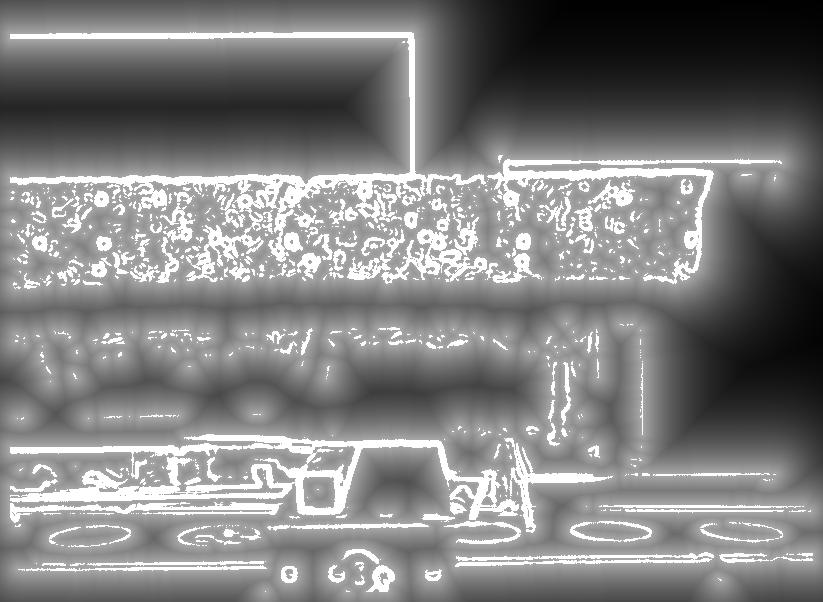}
      \subcaption{Edge score}
       \label{fig:dist_map}
    \end{subfigure}\begin{subfigure}[b]{0.12\textwidth}
      \includegraphics[width=\textwidth]{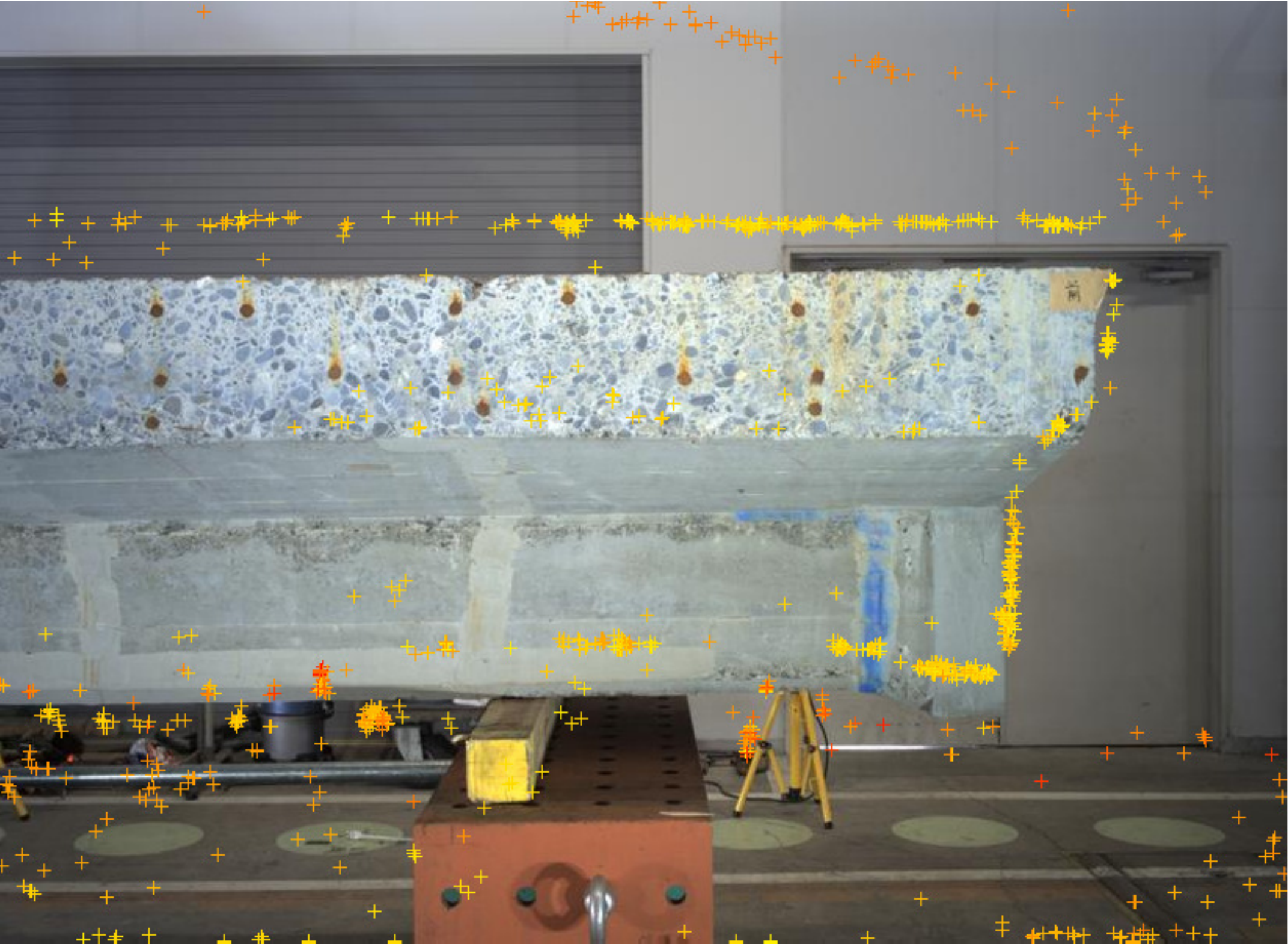}
      \subcaption{Initialization}    
       \label{fig:init_overlay}
    \end{subfigure}\\
    \begin{subfigure}[b]{0.48\textwidth}
      \includegraphics[width=\textwidth]{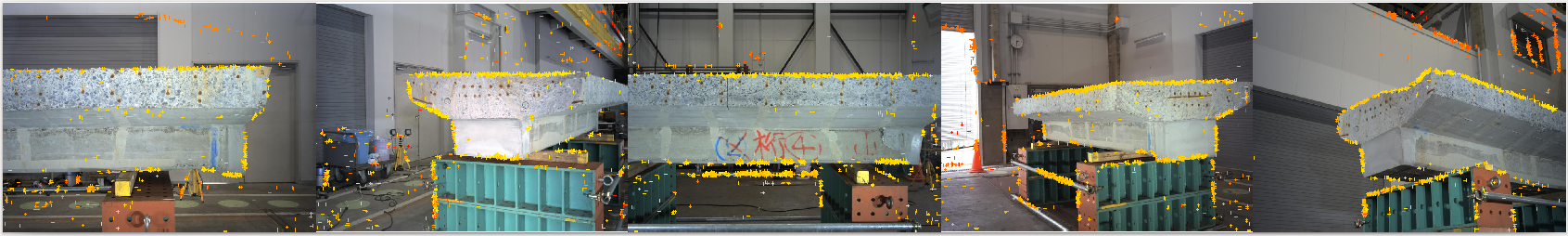}
      \subcaption{Edge alignment after calibration using \cite{levinson2013automatic}}
       \label{fig:calib_proj}
    \end{subfigure}\\
    \begin{subfigure}[b]{0.48\textwidth}
      \includegraphics[width=\textwidth]{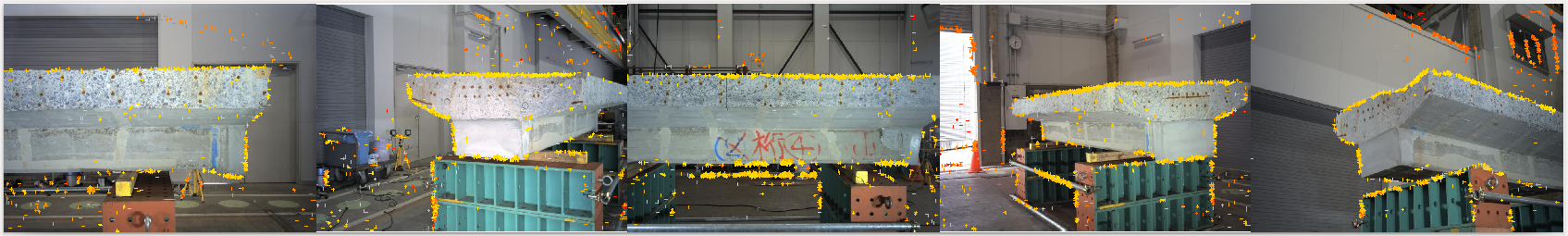}
      \subcaption{Edge alignement after joint optimization}    
       \label{fig:opt_proj}
    \end{subfigure}
\caption{(a)-(d) The key steps of \cite{levinson2013automatic}. (e)-(f) Comparison of extrinsic calibration results from \cite{levinson2013automatic} (e) and ours (f). The color of projected cloud edge points encodes the correlation score: yellow means high while red means low. }\vspace{-3mm}
\label{fig:calibration}
\end{figure}

\blue{In this section,} we evaluate the accuracy of the recovered extrinsic transform. As a comparison, we implemented a target-free calibration method \cite{levinson2013automatic} which uses discontinuities in images and point clouds to iteratively refine an initial guess. The key steps of this method are shown in Fig. \ref{fig:calibration}a-d. Basically, the initial guess is perturbed in each dimension ($x$, $y$, $z$, roll, pitch, yaw) separately and then moved towards the direction that increases the correlation between image edges and projected cloud edges. Eventually, a locally optimal solution can be found if any further changes will decrease the edge correlation. 

\begin{figure} [t]
    \centering
    \includegraphics[width=0.8\linewidth]{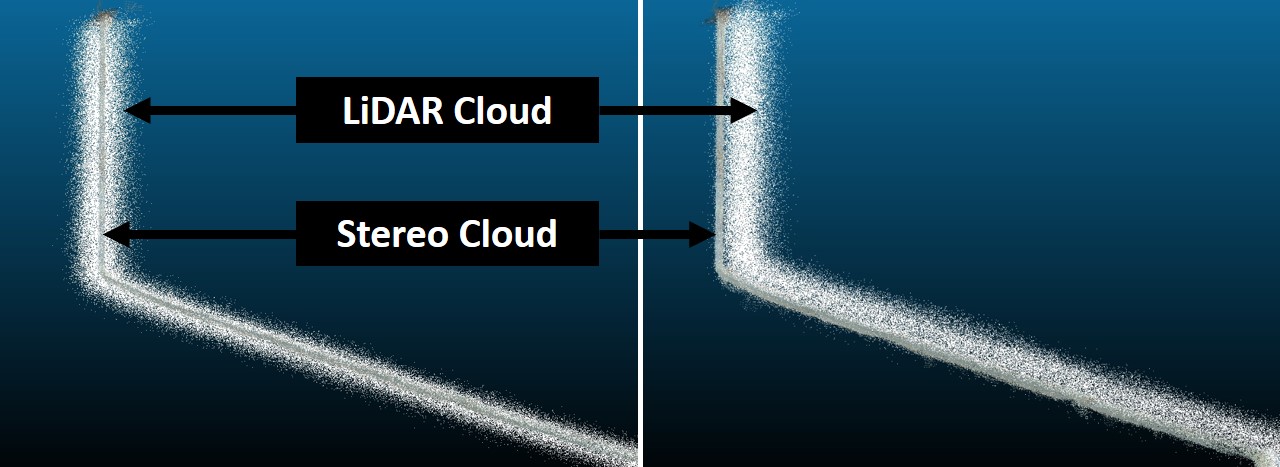}
\caption{Cutaway view of the overlaid LiDAR clouds (white) and stereo clouds (textured). \textit{Left:} Jointly optimized. \textit{Right:} Calibrated using \cite{levinson2013automatic}.} \vspace{-5mm}
\label{fig:alignment_3d}
\end{figure}

Since it is difficult to get ground truth calibration, we choose to compare the extrinsic parameters computed from two methods. The extracted point cloud edges are projected on to the image plane and the projection is visualized in Fig. \ref{fig:calib_proj} and \ref{fig:opt_proj}. However, the edges are both well aligned and no obvious difference can be identified. We then compare the overlay of LiDAR clouds and stereo clouds (see Fig. \ref{fig:alignment_3d}). It can be observed that with our results, the models are aligned consistently while there exists an offset if calibrated using \cite{levinson2013automatic}. Further investigation shows that the offset happens along the camera's optical axis, in which direction the motion will generate less flow on the image. As a result, the total correlation score becomes less sensitive to the motion of the LiDAR along the optical axis. This observation suggests that calibration methods using direct feature alignment, including target-based and target-free, may require wide angle lenses. 

\subsection{Observability of Extrinsic Transform}

The uniqueness conditions stated in Section \ref{approach} basically requires the sensor pod to change its position and orientation for different stations. {In this section, we aim at providing more intuition behind the formal statements. Specifically, the conditions are experimentally demonstrated by perturbing the extrinsic parameters around their optimal values. Three tests are designed to clarify the situations of degeneration.}
\subsubsection{Rotation is fixed}
\begin{figure} [t]
    \centering
    \includegraphics[trim=30 5 30 0,clip,width=0.95\linewidth]{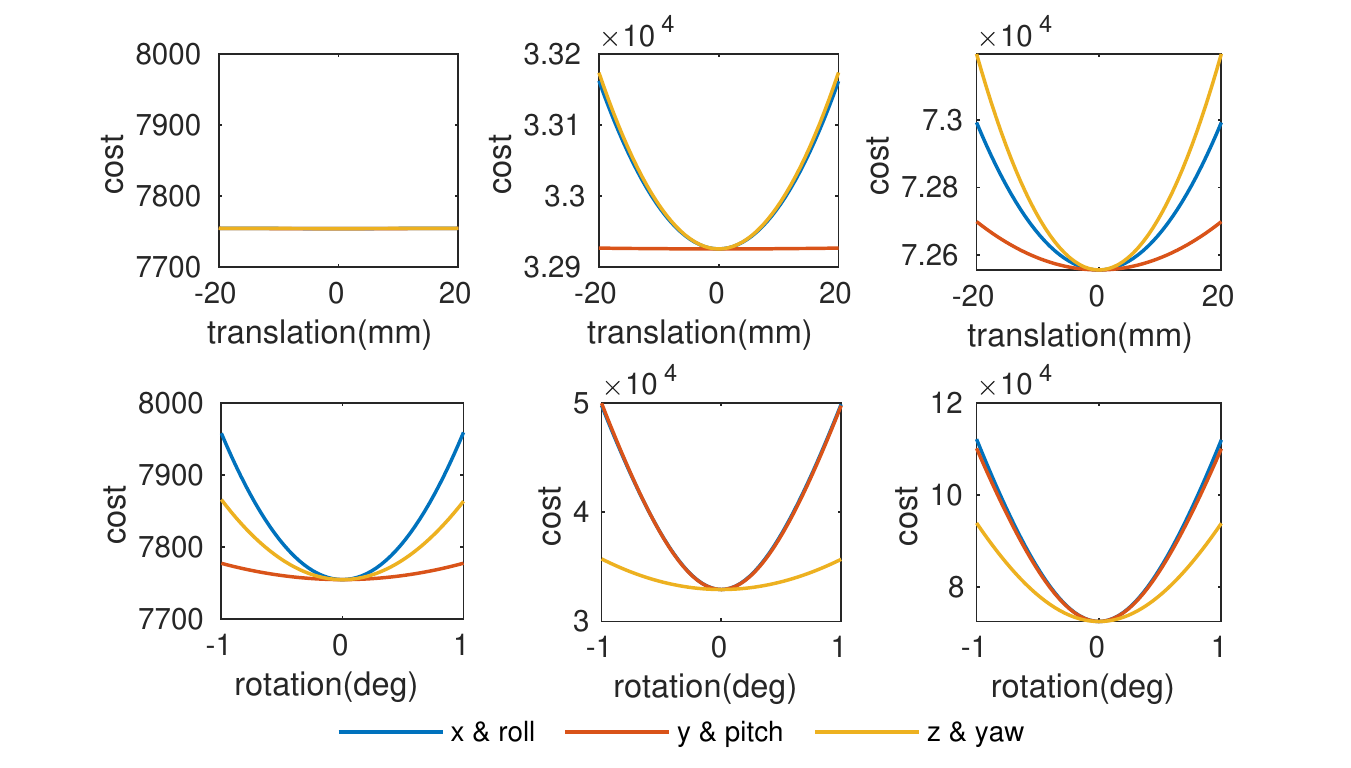}
\caption{\blue{Changes of cost values w.r.t. perturbed extrinsic transform. From left to right, the three columns show the cost changes in three tests: with rotation fixed, with rotation about one axis, and with rotation about two axes. Within each test, translation (top plots) and rotation (bottom plots) perturbations are visualized separately.}}\vspace{-5mm}
\label{fig:obs}
\end{figure}

In this case, the sensor pod is placed at 3 different positions but keeps its orientation unchanged. Specifically, station 1-3 are used for optimization. The total cost after the perturbation is visualized in the left 2 plots of Fig. \ref{fig:obs}. It can be seen that perturbing the translation won't affect the cost value at all, meaning unobservable. Besides, since the 3 frames are almost collinear, the pitch angle is also under-constrained (flat orange curve). 

\subsubsection{Rotation about one axis}
In this case, stations 1-17 are used, where the sensor pod is placed around the T-shaped specimen and all rotations are about the camera's $y$-axis. As shown in the middle plots of Fig. \ref{fig:obs}, position $y$ is under-constrained. 

\subsubsection{Rotation about two axes} For reference, we show the perturbed cost with all 25 available datasets in the right plots of Fig. \ref{fig:obs}. In this case, the rotations can be about $x$- or $y$-axis. As expected, the extrinsic transform is well constrained. 
​​
\subsection{Model Accuracy Evaluation}
\begin{figure} [t]
    \centering
    \includegraphics[width=0.99\linewidth]{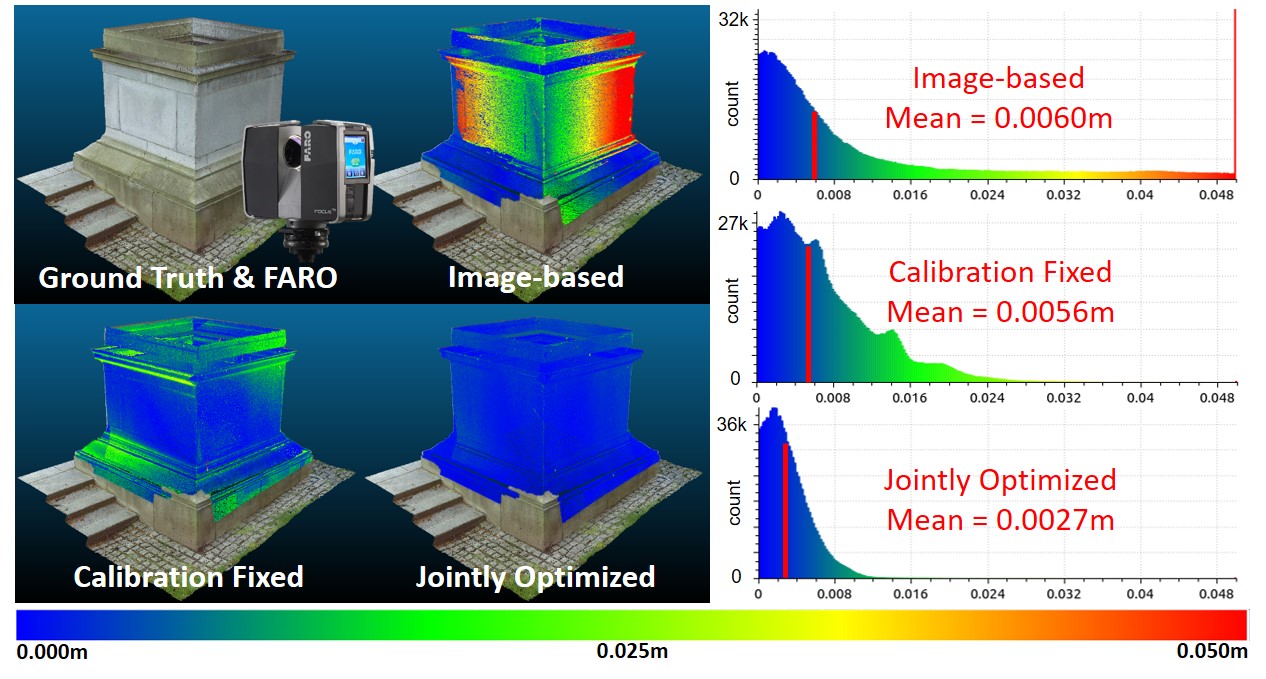}
\caption{\blue{Comparing the reconstructed models with the ground truth model built by the FARO scanner. On the left are visualizations of the ground truth model and the distance map of reconstructed models, where the color encodes the distance error between two point clouds. On the right are the distance histograms corresponding to each comparison and the averaged errors are marked by the red vertical bar.}}\vspace{-3mm}
\label{fig:faro}
\end{figure}

{Since the ground truth data are not available during the test in Tokyo, \blue{we evaluate the reconstruction accuracy on the squared concrete pillar instead.} A FARO $\text{FOCUS}^{\text{3D}}$ scanner (see Fig. \ref{fig:faro}) with $\pm 3$mm range precision is used to obtain the ground truth. The comparison is performed by measuring the point to plane distance between the reconstructed model and the ground truth after precise ICP registration. Furthermore, we compare the results of three models reconstructed using: (1) stereo images only (standard stereo BA), (2) both LiDAR and stereo data but extrinsic calibration is pre-calibrated using \cite{levinson2013automatic}, and (3) both LiDAR and stereo data with extrinsic calibration being adjusted jointly (proposed in this work). \blue{Comparisons (1) and (2) share the same cost function in (3). However, in comparison (1) LiDAR observations are set to have zero weights and $\bold T_e$ is fixed, and in comparison (2) only $\bold T_e$ is fixed during optimization.}}

{The error maps and histograms are visualized in Fig. \ref{fig:faro}. It can be observed that fusing LiDAR data helps to reduce the model error from \blue{6mm to 2.7mm}, which already lies in the precision range of the ground truth. \blue{In fact, due to the limited number of matches between some image frames, the pure image-based model does not align well, resulting in multiple layers of the surface.} Compared with the pre-calibrated case, jointly optimizing the calibration \blue{improves the overall model accuracy and we also benefit from the convenience of self-calibration}. 
Additionally, since our model is reconstructed from multiple sets of data and each station is collected close to the wall (2-3 meters), it measures about 70 points/cm$^2$, which is much denser than the ground truth (10-15 points/cm$^2$). The evaluation results are obtained using the \textit{CloudCompare} software. 
}


\section{Conclusions} 
\label{sec:conclusion}
​This paper presents a joint optimization approach to fuse LiDAR and camera for pose estimation and dense reconstruction. It is shown to be able to build dense 3D models and recover camera-LiDAR extrinsic transform accurately. Besides, the accuracy of the reconstructed model is evaluated by comparing to a ground truth model and it shows our method can achieve accuracy similar to a survey scanner.

The proposed method requires data to be collected station by station, which can be time consuming and inconvenient if the viewpoint is difficult to access. For example, the I-shaped beams supporting the deck of a bridge are usually too high to reach. Therefore, future work will be focused on handling sequential data with the sensor pod moving in the environment. Micro Aerial Vehicles (MAVs) may also be used to carry the sensor pod. \blue{Another thread of future work is to improve the quality of stereo reconstruction. For instance, given the LiDAR-camera extrinsic calibration obtained from our method, probabilistic fusion methods such as \cite{maddern2016real} can be applied to recover a dense local map.}
\section{Acknowledge}
This work is supported by the Shimizu Institute of Technology, Tokyo. The authors are grateful to Daisuke Hayashi for his help with experiments in Japan. We also thank Huai Yu, Hengrui Zhang and Ruixuan Liu for building the sensor pod and helping with data collection.

\bibliographystyle{IEEEtran.bst}

\end{document}